# Conveying Surroundings Information of a Robot End-Effector by Adjusting Controller Button Stiffness

Noel Alejandro Avila Campos[1], Masashi Konyo[1], Shotaro Kojima[1], Ranulfo Bezerra Platurco[1], and Satoshi Tadokoro[1]

[1] *Human-Robot Informatics. Tohoku University, Sendai, Japan*
(Email: avila.noel@rm.is.tohoku.ac.jp)

**Abstract ---** This study addresses the challenge of low dexterity in teleoperation tasks caused by limited sensory feedback and visual occlusion. We propose a novel approach that integrates haptic feedback into teleoperation using the adaptive triggers of a commercially available DualSense controller. By adjusting button stiffness based on the proximity of objects to the robot's end effector, the system provides intuitive, real-time feedback to the operator. To achieve this, the effective volume of the end effector is virtually expanded, allowing the system to predict interactions by calculating overlap with nearby objects. This predictive capability is independent of the user's intent or the robot's speed, enhancing the operator's situational awareness without requiring complex pre-programmed behaviors. The stiffness of the adaptive triggers is adjusted in proportion to this overlapping volume, effectively conveying spatial proximity and movement cues through an "one degree of freedom" haptic feedback mechanism. Compared to existing solutions, this method reduces hardware requirements and computational complexity by using a geometric simplification approach, enabling efficient operation with minimal processing demands. Simulation results demonstrate that the proposed system reduces collision risk and improves user performance, offering an intuitive, precise, and safe teleoperation experience despite real-world uncertainties and communication delays.

## 1 INTRODUCTION

Teleoperation tasks involve robots being controlled remotely by humans. They are necessary across numerous domains, including nuclear, underwater, space exploration, tele-surgery, industrial, and safety and rescue operations [1] [2]. Efficient execution of teleoperation tasks requires a degree of dexterity depending on the specific activity. Dexterity has been a topic of discussion in robotics [3]. In this document, we define it closely to manual dexterity: the ability to manipulate objects using the hands and fingers to perform movements with precision and in a timely manner for a specific task [4]. These movements must be precise, fully controlled, and responsive to the surrounding environment.

The low dexterity movements of a robot can result in multiple direct damage to the robotic system itself, the environment in which it operates, or the objects it interacts with. If the robot lacks a proprioceptive system, unnoticed collision will increase the difference between the estimated and actual positions.

One reason for the low dexterity in teleoperation is that the teleoperator cannot feel what the robot experiences. Additionally, since the teleoperator and the robot are not in the same environment, the operator relies only on the information the robot provides. When attempting to interact with an object, the robot arm often obstructs visual information, a phenomenon known as occlusion. This visual information is also insufficient to understand the depth of the object. Solutions such as adding cameras increase the number of complex stimuli the user receives, raise the mental load, and result in poorer performance [5].

## 2 PROPOSAL

This study addresses a critical aspect of teleoperation: enhancing the user's awareness of the robot's interaction with the environment, particularly emphasizing its target object. With this purpose in mind, it is proposed to integrate haptic feedback corresponding to the relationship between the robot's end effector and nearby objects. By incorporating haptic feedback, the operator is expected to gain a more intuitive understanding of the environment and improve their control over the robot's interactions.

The proposed technology anticipates interactions by virtually enlarging the volume of the robot's gripper and calculating its volume overlaps with the environment. The same input method is used for haptic feedback to convey this information intuitively and in real time to the user. The generated stimulus consists of changing the stiffness of the buttons on the robot's controller, which corresponds to the direction of the detected objects nearby.

## 3 IMPLEMENTATION

### 3.1 Simulation

We rely on a simulation of the robot and its surroundings to get information from the environment, which can be achieved with depth cameras or radars and segmentation models.

Given the error in the estimation of the position of a camera by the device's specification:

$$Cam_{Err} = x_{real} - \hat{x}_{est} \tag{1}$$

Assuming it behaves as a normal distribution:

$$Cam_{Err} \sim N(\mu_c, \sigma_c^2) = \frac{1}{\sigma\sqrt{2\pi}} exp(-\frac{(x-\mu_c)^2}{2\sigma_c^2}) \quad (2)$$

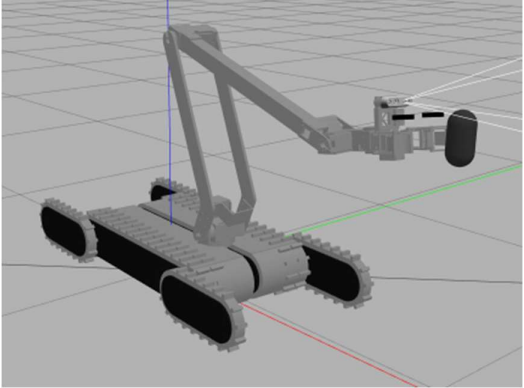

Fig.1 Example of simulation of the robot and environment

### 3.2 End-Effector Position estimation

As the state of the end-effector is essential, a precise estimation of the current position and orientation of the end-effector can be based on the joint angle information. Additionally, the distribution of the position error of the end-effector can be estimated through error propagation.

Given: $\theta_i$ = angle of the joint $i$-th.

The difference between the actual and estimated angles can be quantified by the error specified in the encoder's characteristics.

$$\triangle \theta_i = \theta_i - \widehat{\theta_i} \quad (3)$$

Given the direct kinematics using a series of transformation matrices:

$$T_n = T_1 \cdot T_2 \cdots T_{n-1} \cdot T_n \quad (4)$$

Where:
$T_i$ = The transformation matrix for joint i

The propagation of errors is [6]:

$$\sigma_p = \sqrt{\sum_{i=1}^{n}\left(\frac{\partial p}{\partial \theta_i} \triangle \theta_i\right)^2} \quad (5)$$

Where:
$\sigma_p$: The standard deviation of the error in the end-effector's position estimation.
$\frac{\partial p}{\partial \theta_i}$: Derivative of position with respect to $i$-th angle.

The error of the end effector position estimation by direct kinematic would correspond to:

$$\triangle T_{ee,k} \equiv T_{ee} - \widehat{T_{ee,k}} \quad (6)$$

Assuming it as a normal distribution, its model is:

$$\triangle T_{ee,k} \sim N(\mu_{ee}, \sigma_p^2) \quad (7)$$

### 3.3 Delay displacement consideration

We should also consider the unavoidable delay in teleoperation to correctly calculate the difference in position between the robot and its virtual counterpart.

$$\triangle T_{ee,d} \equiv \frac{dx}{dt} \cdot t_d \quad (8)$$

### 3.4 Enlargement Distance

Then, we enlarge the virtual model of the detected objects and environment. The enlargement's distance should be big enough to overcome the estimated position error, the discrepancy of the virtual object with its physical counterpart, and the delay in communication to prevent any possible collision.

Nevertheless, the distance should be small so as not to slow down the whole performance of the robot more than convenient. Another parameter to consider is the distance between both sides of the robot's gripper; otherwise, the enlarged volume could cause interference between them. As the contact depends on the relation of both objects, all considered errors should be considered. We combine normal distributions for the camera and end effector position estimation, equations (2), (7), and (8). We applied the following mathematical approach to set the distance for enlargement. Based on the safety parameter P, which represents the probability requirement for preventing undesired contact, the enlargement distance is defined by:

$$D_{incr}(P) \approx \triangle T_{ee,d} + (\triangle T_{ee,k})^{-1} + (Cam_{Er})^{-1} \quad (9)$$

$$D_{incr}(P) \approx \triangle T_{ee,d} + \Phi^{-1}(P, \mu_{ee} + \mu_c +, \triangle p^2 + \sigma_c^2)(10)$$

Where:
$P$ = Probability for preventing collision ($\in [0,1]$)
$\Phi$ = Cumulative distribution function until P [12].

$$\Phi(P, \mu, \sigma^2) = \int_{-\infty}^{P} \frac{1}{\sigma\sqrt{2\pi}} exp(-\frac{(t-\mu)^2}{2\sigma^2})dt$$

It is worth mentioning that the enlargement distance is calculated only once based on the robot to be implemented based on the sensors' information.

## 4 EXECUTION

### 4.1 Volume intersection calculation

As the objects in the simulation can be modeled as basic shapes, the volume intersection in three dimensions can be calculated by applying analytical methods.

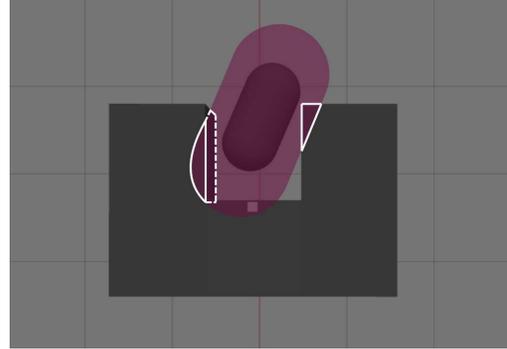

Fig.2 Enlarged volume overlap representation

Defining the recognized objects as spheres and tetrahedrons on the simulation, the intersection between a Tethraedro and a sphere.

$$V_{overlap} = \int_{(x,y,z)\in Tet} H(R^2 - \{(x-x_0)^2 + (y-y_0)^2 + (z-z_0)^2\})dxdydz \quad (11)$$

Where:
$H(x)$ = The Heaviside function.
- $H(x) = 1 \quad for\ x \geq 0$
- $H(x) = 0 \quad for\ x < 0$

$R$ = Sphere's radio
$(x_0, y_0, z_0)$ = Coordinates of Sphere's center

$Tet$ = Tethraedro's Coordinates

As the objects are simplified as spheres and boxes, we can lower the algorithm's complexity by applying the following approximation [8].

$$V_{overlap} = V_s - \sum_{faces}^{\square} V_c + \sum_{edges}^{\square} V_w - \sum_{vertices}^{\square} V_{co} \quad (12)$$

Where it represents the Volume of:

s = sphere, c = cap, w = wedge, co = cone

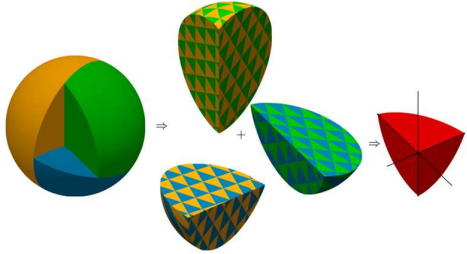

Fig.3   Sphere's volume overlap calculation [8]

### 4.2   Haptic Feedback

The PlayStation 5 DualSense controller was used as it features adaptive triggers on the "R2" and "L2" buttons. These triggers have an internal mechanical system that can adjust the stiffness of the buttons, providing different force reactions. The stiffness of the controller's buttons is adjusted proportionally to the volume intersection, allowing the user to receive haptic feedback about the objects surrounding the robot's end effector (see Fig. 4).

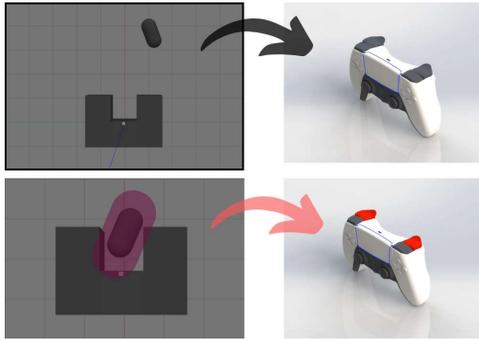

Fig.4   Adjustment of the button stiffness

### 4.3   Visual Feedback

The visual environment was presented to the user in two perspectives. The first one is from above, with a broader view. Another perspective of visual feedback is in the first person. The second is closer to a real case in which the end-effector carries a camera (see Fig. 5).

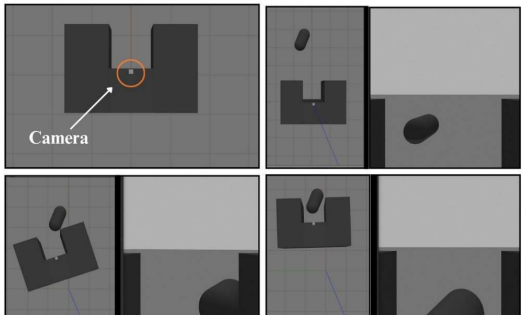

Fig.5   Visual feedback examples

## 5   CONCLUSION

As a result, the system prevents any potential contact that the end-effector may have. It successfully alerts the user, enabling them to make informed decisions when interacting with the environment. Haptic feedback exerts resistance against the user's input, leading to slower and more controlled movements when the gripper is close to another object. Additionally, it leads to a better gripping position on the object and helps the user understand the target's depth distance.

Compared to previous solutions, the proposal has few hardware requirements and proposes a low computational load calculation method because of the geometrical simplification approach. The adjustment in the stiffness of the button related to the robot's movement on a commercial controller result in a "one degree of freedom" haptic feedback. Additionally, this method allows intuitive transmission of the information to the user.

This solution is robust because it remains effective despite differences between real-world conditions and the simulation environment. It also works if the estimated position of the end-effector contains errors. Even with transmission delays, it helps prevent collisions. The method reduces computational complexity by focusing on nearby modeled objects instead of complex meshes.

### REFERENCES

[1]   Application of Adaptive Controllers in Teleoperation Systems: A Survey. (2014). IEEE Transactions on Human-Machine Systems, 44(3), 337–352. doi:10.1109/thms.2014.2303983

[2]   Jean Vertut and Philippe Coiffet. 1986. Teleoperations and robotics: evolution and development. Prentice-Hall, Inc., USA.

[3]   R. R. Ma and A. M. Dollar. On dexterity and dexterous manipulation. In 15th International Conference on Advanced Robotics: New Boundaries for Robotics, ICAR 2011, Tallinn, Estonia, June 20-23, 2011., pages 1–7, 2011.

[4]   Makofske, B., & Power, E. (2018). Manual Dexterity. Encyclopedia of Clinical Neuropsychology, 2080–2081. doi:10.1007/978-3-319-57111-9_1460

[5]   Darvish, K., et al. (2023). Teleoperation of Humanoid Robots: A Survey.10.48550/arXiv:2301.04317.

[6]   Siciliano, Bruno & Sciavicco, L. & Luigi, Villani & Oriolo, G.. (2009). Robotics: Modelling, planning and control.

[7]   Walpole, Ronald E., et al. (2016). Probability & Statistics for Engineers & Scientists (9th ed., Global ed.). Tokyo: Pearson.

[8]   Strobl, S., Formella, A., & Pöschel, T. (2016). Exact calculation of the overlap volume of spheres and mesh elements. Journal of Computational Physics, 311, 158-172. https://doi.org/10.1016/j.jcp.2016.02.003